\documentclass[journal]{IEEEtran}
\IEEEoverridecommandlockouts
\usepackage{cite}
\usepackage{tikz}
\usetikzlibrary{shapes.geometric, arrows}
\usepackage{url}
\usepackage{soul}
\usepackage{amsmath,amssymb,amsfonts}
\usepackage{algorithmic}
\usepackage{textcomp}
\usepackage{graphicx}
\usepackage{xcolor}
\usepackage{multirow}
\usepackage[detect-all]{siunitx}
\def\BibTeX{{\rm B\kern-.05em{\sc i\kern-.025em b}\kern-.08em
    T\kern-.1667em\lower.7ex\hbox{E}\kern-.125emX}}

\tikzstyle{io} = [rectangle, minimum width=3cm, minimum height=1cm, text centered, draw=black]
\tikzstyle{convolution} = [rectangle, minimum width=3cm, minimum height=1cm, text centered, draw=black]
\tikzstyle{normalize} = [rectangle, minimum width=3cm, minimum height=1cm, text centered, draw=black]
\tikzstyle{relu} = [rectangle, minimum width=3cm, minimum height=1cm, text centered, draw=black]
\tikzstyle{pooling} = [rectangle, minimum width=3cm, minimum height=1cm, text centered, draw=black]
\tikzstyle{dense} = [rectangle, minimum width=3cm, minimum height=1cm, text centered, draw=black]
\tikzstyle{soft} = [rectangle, minimum width=3cm, minimum height=1cm, text centered, draw=black]
\tikzstyle{class} = [rectangle, minimum width=3cm, minimum height=1cm, text centered, draw=black]
\tikzstyle{arrow} = [thick,->,>=stealth]

\graphicspath{{figures/}}

\begin{document}

\title{A Deep Learning Approach to Tongue Detection for Pediatric Population}

\author{Javad Rahimipour Anaraki, Silvia Orlandi, \IEEEmembership{Member, IEEE}, and Tom Chau, \IEEEmembership{Member, IEEE}
\thanks{J. R. A. and T. C. are with the Institute of Biomedical Engineering, University of Toronto, Rosebrugh Building, 164 College Street, Toronto, Ontario M5S 3G9 Canada (e-mail: j.rahimipour@utoronto.ca and tom.chau@utoronto.ca). J. R. A., S. O. and T. C. are with the Bloorview Research Institute, Holland Bloorview Kids Rehabilitation Hospital, 150 Kilgour Road Toronto, Ontario M4G 1R8 Canada (e-mail: sorlandi@hollandbloorview.ca).}}

\markboth{}
{Shell \MakeLowercase{\textit{et al.}}: Bare Demo of IEEEtran.cls for IEEE Journals}
\maketitle

\begin{abstract}
Children with severe disabilities and complex communication needs face limitations in the usage of access technology (AT) devices. Conventional ATs (e.g., mechanical switches) can be insufficient for nonverbal children and those with limited voluntary motion control. Automatic techniques for the detection of tongue gestures represent a promising pathway. Previous studies have shown the robustness of tongue detection algorithms on adult participants, but further research is needed to use these methods with children. In this study, a network architecture for tongue-out gesture recognition was implemented and evaluated on videos recorded in a naturalistic setting when children were playing a video-game. A cascade object detector algorithm was used to detect the participants' faces, and an automated classification scheme for tongue gesture detection was developed using a convolutional neural network (CNN). In evaluation experiments conducted, the network was trained using adults and children's images. The network classification accuracy was evaluated using leave-one-subject-out cross-validation. Preliminary classification results obtained from the analysis of videos of five typically developing children showed an accuracy of up to 99\% in predicting tongue-out gestures. Moreover, we demonstrated that using only children data for training the classifier yielded better performance than adult's one supporting the need for pediatric tongue gesture datasets. 
\end{abstract}

\begin{IEEEkeywords}
Tongue Detection, Deep-learning Techniques, Children, Assistive Technology
\end{IEEEkeywords}

\section{Introduction}
The worldwide prevalence of children with communication and motor disabilities is difficult to estimate. The Global Burden of Disease reported that 93 million children aged 0-14 years experience moderate or severe disability \cite{WHO}. An individual with complex communication needs (CCN) might not be able to communicate through regular means (e.g., speech or writing). Alternative pathways to communication include gestures, facial expressions, vocalizations, and low- or high-tech devices. Individuals with CCN are usually those suffering from congenital disabilities (e.g., cerebral palsy (CP); autism spectrum disorder; developmental disabilities; intellectual disability) and acquired disabilities (e.g., traumatic or acquired brain injuries; temporary conditions for patients in critical care settings). For example, children with CP, the most common cause of physical disability in childhood that affects 2 per 1,000 live births \cite{oskoui2013update, CDCdata}, may have difficulty manipulating conventional interface devices due to limited voluntary muscle control and reduced muscle tone \cite{chau2012home}. Moreover, children with CP usually present CCN. In fact, one in four children with CP is not able to talk \cite{novak2012clinical}, and 80\% of people with CP have speech impairments \cite{odding2006epidemiology}. Besides, more than 60\% of children with bilateral CP have limited manual ability \cite{arner2008hand}. 

The use of augmentative and alternative communication (AAC) devices can support them in developing alternative means of communication. The prevalence of AAC users is difficult to estimate due to the wide variability across the individuals that require the use of these devices based on the diagnosis, age, geographical position, communication modality. According to the National Survey of Children With Special Health Care Needs, approximately 2.1\% of children and youth with disabilities require AAC systems. Still, the need for communication aids and AAC devices is unmet for approximately 25\% of these children \cite{dusing2004unmet}, and the difficulty in physical access due to limited motor control is one of the main reasons for the abandonment of access technology (AT) \cite{chau2012home, waller2019telling}. As such, individuals with CCN may require customized AT devices to help improve participation in educational and play activities \cite{waller2019telling}. An AT is a form of assistive technology whose key purpose is to translate a user's intent into a control signal for a user interface and eventually into a functional activity. However, conventional motor-based pathways (e.g., mechanical switches) have limited viability for children with CCN who can have difficulty controlling the force required to activate a switch, limited ability to release a switch once activated, and a tendency to activate a switch multiple times \cite{chau2012home, waller2019telling, lancioni2011assistive, metaxas2013review}. Previous studies explored alternative solutions to develop customized ATs for children with CCN through signal processing and machine learning techniques using facial gestures \cite{myrden2014trends}, dysarthric speech and vocalizations \cite{orlandi2019audio, falk2010augmentative, lui2012development}, head movements \cite{orlandi2020head, betke2002camera}, eye-movements \cite{karamchandani2015development, karlsson2018eye}, and brain signals \cite{weyand2017challenges}. One of the alternative pathways for interaction in individuals with CCN is through the ability to control tongue protrusion and movements \cite{leung2010multiple}. In fact, nonspeaking individuals may use their tongue as a means of communication. The use of tongue gestures as tongue-operated AT devices has already experimented with adult able-bodied populations \cite{li2015tongue, goel2015tongue, nguyen2018tyth, hashimoto2018, sapaico2007toward, Viola01rapidobject, caltech, sapaico2011detection, fox2005valid, sapaico2011visual, niu2019tongue, van2007edge, dalka2014visual}. Furthermore, ATs tailored for children with CCN are not common \cite{pousada2011use}, and methods developed for adults do not always work successfully with young children \cite{srinivas2019face}. 
Various studies applied video-based approaches and machine learning algorithms to detect tongue gestures for AT devices \cite{leung2010multiple, sapaico2007toward, sapaico2011detection, sapaico2011visual, niu2019tongue, kim2010wireless, xin2018automatic, wang2020artificial}. Deep learning approaches and convolutional neural network (CNN) have already been utilized to detect tongue during speech analysis \cite{wang2016optimal, kroos2017using, wu2018predicting}. To the best of our knowledge, CNNs have not been applied to colour images for tongue gesture detection. Furthermore, the application of machine learning and deep learning techniques on AT based on facial gestures tailored for children is also limited by the lack of accessible datasets of images and videos of pediatric populations \cite{srinivas2019face, dapogny2019automatically}. If children face databases are limited \cite{srinivas2019face}, tongue gesture datasets on pediatric age are just not available. In the literature, there are two datasets of tongue gestures, but they include only images of adults \cite{DVN/COJZMQ_2019, SP2/5T2RD9_2019}. So far, all of the reported methods and systems have been designed and trained against datasets collected from adults. Facial features changes and developments from childhood to adulthood has made the already proposed methods relatively unsuitable for the applications where children are involved, particularly when they have any form of disability.

In this paper, we investigate the possibility of using adults' tongue gestures to generate a model and test it against children's tongue movements. We also present a dataset of facial and tongue gestures that we collected from five pediatric participants to train a model specifically for children. The remainder of the paper is organized as follows: In Section \ref{sec:work}, we provide an overview of the previous works on AT devices based on tongue detection and video-based approaches for the analysis of tongue gestures. In Section \ref{sec:proposed}, we report on the dataset collected from the five children with four labels, namely, neutral, tongue-out, smiling and mouth opening, and video annotation method. Then we propose a CNN architecture to train on the collected data to detect tongue-out gestures with reasonable accuracy in children. The results are presented in Section \ref{sec:results}, followed by a discussion in Section \ref{sec:discussion} where we investigate the need for collecting data from children to provide more accurate and reliable models using neural network-based methods. Finally, we conclude our paper in Section \ref{sec:conclude}.

\section{Related Work}\label{sec:work}
Several attempts have been made to develop methods and devices for tongue gesture detection with the aim to support people with severe disabilities \cite{li2015tongue, goel2015tongue, nguyen2018tyth, hashimoto2018, sapaico2007toward, sapaico2011detection, niu2019tongue}. Most of the studies to date have focused on the development of contactless sensors \cite{li2015tongue, goel2015tongue, nguyen2018tyth, hashimoto2018}. Li et al. \cite{li2015tongue} developed a non-invasive and contact-less micro-radar sensor called Tongue-n-Cheek sat on an acrylic sheet attached to a helmet to detect tongue movements which were later analyzed to extract tongue gestures. Based on the designed system, they were able to recognize six tongue gestures for steering a powered wheelchair with 95.00\% of accuracy on a total of 420 gesture repetitions performed by five able-bodied adults. One drawback of the proposed system was the portability and the convenience of using a helmet at all the time, which can be troublesome for some individuals. 
Goel et al. \cite{goel2015tongue} implemented a wireless non-invasive and non-contact device using three motion sensors, two located at each side of the face and one at the mouth front to detect tongue motion, cheek puffing, and jaw movements. They evaluated the system against the data collected from eight participants and achieved 94.30\% of accuracy using SVM classifier. 
Nguyen et al. \cite{nguyen2018tyth} proposed a system called typing on your teeth (TYTH) which was composed of electroencephalography (EEG), electrocardiogram (EKG), and skin surface deformation (SKD) sensors placed behind the ears to capture brain and muscle activities and skin deformation signals. TYTH was able to recognize different tongue movements using wavelet-coefficient analysis, Gaussian mixture model, and support vector machine (SVM) classifier. They tested the system against the collected data from 15 able-bodied adults, reaching 88.61\% of accuracy. However, even in this case, the requirement of wearing a device to be able to use a tongue gesture detection system could be prohibitive. Hashimoto et al. \cite{hashimoto2018} introduced a hardware that used an array of photo-reflective sensors placed in a mouthpiece to actively measure the distance of the tongue from the surroundings. The proposed system was able to detect four tongue positions with 85.67\% of accuracy. One of the main limitations of the proposed system was the requirement of a customized mouthpiece for every individual.

In an attempt to translate tongue gestures (i.e., left and right movements) to emulate mouse clicks, Sapaico and Nakajima designed a system to detect mouth region using a three-layer cascade SVM classifier \cite{sapaico2007toward}, and a pre-process stage where they used Viola and Jones face detection algorithm \cite{Viola01rapidobject}. In the first layer, the system decides whether the tongue is present or not and pass the results to the second layer. If the tongue is present, the second layer decides whether the tongue position is in the middle. If not, the last layer determines if tongue is located either on the left or right side of the mouth. They tested their system against the data collected from 10 adult participants combined with Caltech frontal face dataset \cite{caltech}. Their system achieved 89.80\%, 85.15\% and 88.46\% of success rate for layers one, two, and three, respectively. Although they created a balanced dataset, the number of samples in the training set was small (i.e. 312 samples for a three-class dataset). 

In another work, Sapaico et al. \cite{sapaico2011detection} introduced a new method to detect mouth region by applying Gabor filters to improve mouth features, Hough line detector to find mouth corners, and eventually a bounding box for the mouth region. The region was then used to create templates for mouth-close/tongue-out, tongue-left and tongue-right gestures using each half, and detected gestures using normalized correlation matching. The proposed system achieved 90.20\% of accuracy for tongue-out gestures and 84.78\% of accuracy for the left and right ones against 300 images adopted from \cite{fox2005valid, caltech}, \cite{sapaico2011detection} and a collection gathered by the authors. Also, Sapaico et al. \cite{sapaico2011visual} used the proposed system to translate mouth gestures to Morse code for visual text entry. 

In a recent work by Niu et al. \cite{niu2019tongue}, a two-phase tongue gesture recognition method to classify six tongue and mouth gestures, namely, mouth-close, mouth-open, tongue-up, tongue-down, tongue-left and tongue-right, was proposed. First, a detected face was processed using an edge enhancement algorithm \cite{van2007edge} and then passed to a lip contour detection method \cite{dalka2014visual}. Next, the results were used to train six classifiers using AdaBoost algorithm to detect each gesture. In the first phase, their method's average classification accuracy was 83.32\% for an online direction selection task using four gestures against 1,800 images collected from six adult participants. In a second application of moving the cursor to left and right using two tongue gestures, the proposed method achieved 84.60\% of accuracy. In the second phase, they provided a mirror-view of the participant on the screen and a visual tongue gesture indicator to improve the participant's performance. However, they reported no significant improvements in the results.

Although several studies \cite{goel2015tongue, nguyen2018tyth, hashimoto2018, ostadabbas2016tongue, huo2007magnetic, struijk2006inductive, cheng2014tip, kim2010wireless} investigated the use of tongue sensor-based techniques to support the interaction with another device (e.g., laptop or smartphone), the use of computer vision-based \cite{leung2010multiple, sapaico2007toward, sapaico2011detection, sapaico2011visual, niu2019tongue, kim2010wireless} and artificial intelligence \cite{sapaico2007toward, sapaico2011detection, niu2019tongue, xin2018automatic, wang2020artificial} techniques remains a field to be explored for developing user-friendly AT devices. Moreover, the development of non-invasive and contact-less ATs can be beneficial for people with CCN to reduce the risk of abandonment \cite{sugawara2018abandonment, federici2016abandonment}. However, only two studies tested their algorithms on people with disabilities \cite{leung2010multiple, ostadabbas2016tongue} and only one \cite{leung2010multiple} investigated the use of video-based techniques for tongue detection in children with disabilities. Leung and Chau \cite{leung2010multiple} developed a video-based AT for tongue protrusion recognition for a 7-year-old child with severe spastic quadriplegic CP. Their system used three cameras and was tested in a controlled environment reaching an average sensitivity of 82.00\% and specificity of 80.00\%.

\section{Proposed Approach}\label{sec:proposed}

\subsection{Dataset}
In this section, an adult dataset and the one collected from children are introduced.

\subsubsection{Adults' dataset} We used the RGB-D tongue dataset from \cite{SP2/5T2RD9_2019}, which contains 17 adult participants (P01-P17) performing seven gestures, mouth opening, mouth closing, tongue-up, tongue-down, tongue-middle, tongue-left, and tongue-right. Out of all seven classes, we only used mouth closing (i.e. neutral face) and tongue-down (i.e. tongue protrusion) to match our collected dataset from children. It is worth noting that the RGB-D tongue dataset only includes images of individuals from nose to chin, and no gender and age information was provided.

\begin{table}[h]
\caption{The number of images in each class for each participant in the RGB-D tongue dataset}
\centering
    \begin{tabular}{ c c c }
     \textbf{Participant} & \textbf{\#Neutral face} & \textbf{\#Tongue-out} \\\hline
     P01 & 237 & 230 \\
     P02 & 165 & 137 \\
     P03 & 223 & 230 \\
     P04 & 204 & 210 \\
     P05 & 232 & 199 \\
     P06 & 246 & 277 \\
     P07 & 235 & 282 \\
     P08 & 237 & 285 \\
     P09 & 253 & 243 \\
     P10 & 243 & 281 \\
     P11 & 209 & 260 \\
     P12 & 235 & 267 \\
     P13 & 213 & 257 \\
     P14 & 245 & 293 \\
     P15 & 244 & 269 \\
     P16 & 218 & 252 \\
     P17 & 228 & 287 \\\hline
     \textbf{Total} & 3867 & 4259
    \end{tabular}
\label{adultsData}
\end{table}

\subsubsection{Children's dataset} Five typically developing children aged 6 to 18 were recruited at Holland Bloorview Kids Rehabilitation Hospital (Toronto, Canada). Table \ref{kidsData} reports gender, age, and the number of frames for each class (i.e., neutral face and tongue protrusion) used for each participant (C01-C05). Participants were invited to come in for a session where they were asked to repeat mouth and tongue gestures (i.e., tongue protrusion, smiling and opening mouth) several times, through an interaction with a custom video game developed with \mbox{Unity 2019.2.0f1}\cite{orlandi2020head}. The main character of the video game had to run across several platforms. At the end of each platform, a prompt with the word “jump” appeared on the screen and the child had to perform the movements. The researcher then pressed the keyboard space-bar to simulate the action for the participant, and the character jumped on the following platform. Each child was recorded using an Intel\textsuperscript{\textregistered} RealSense\textsuperscript{\texttrademark} D415 camera with a frame rate of 30 frames per second, placed 45cm away from the face. All videos were acquired using a frontal view, with controlled lighting conditions. The Research Ethics Board at the Bloorview Research Institute approved this study. Informed written consent was obtained from all participants and their parents, if required.

\begin{table}[h]
\caption{The number of images in each class for each participant in children dataset}
\centering
\addtolength{\tabcolsep}{-2.8pt}
    \begin{tabular}{c c c c c}
     \textbf{Participant} & \textbf{Gender} & \textbf{Age (years)} & \textbf{\#Neutral face} & \textbf{\#Tongue-out} \\\hline
     C01 & M & 17 & 3118 & 587 \\  
     C02 & F & 9 & 3782 & 488 \\  
     C03 & M & 6 & 4096 & 450 \\  
     C04 & F & 6 & 4090 & 338 \\  
     C05 & F & 6 & 3133 & 740 \\\hline
     \textbf{Total} & & & 18219 & 2603
    \end{tabular}
\label{kidsData}
\end{table}

For training and testing, we employed leave-one-subject-out cross-validation method, where 15\% of the training data was user for validation.

\subsection{Video annotation}
Each video's colour stream was manually annotated by a research assistant to generate the ground truth of the facial gesture classes. A user-friendly interface to perform the annotation was developed and implemented in MATLAB\textsuperscript{\textregistered} 9.7.0.1247435 (R2019b), and the code is available on GitHub\footnote{https://hollandbloorview.flintbox.com/\#technologies/a4986f8c-426d-47d1-8865-7284bf389734}. The interface allows uploading AVI videos and playing the video recordings. When an event occurs, the user can mark the beginning and the ending of the event (e.g., tongue protrusion) by pushing a button on the interface. The annotation tool saves frames and times when the event occurred. A total of 15 videos containing 184 tongue gesture repetitions, 163 smile repetitions, and 189 mouth opening repetitions were identified as reported in Table \ref{kidsAnnotationData}. For each gesture, the number of samples is also included in parenthesis.

\begin{table}[h]
\caption{The number of gesture repetitions in each class for each participant in the children dataset. The number of samples for each gesture is presented in parenthesis.}
\centering
    \begin{tabular}{ c c c c }
     \textbf{Participant} & \textbf{\#Tongue-out} & \textbf{\#Smiling} & \textbf{\#Mouth opening} \\\hline
     C01 & 34 (587) & 35 (492) & 31 (509) \\  
     C02 & 34 (488) & 32 (858) & 36 (563) \\  
     C03 & 38 (450) & 36 (2191) & 50 (627) \\  
     C04 & 35 (338) & 31 (1295) & 34 (472) \\  
     C05 & 43 (740) & 29 (1686) & 38 (675) \\\hline
     \textbf{Total} & 184 (2603) & 163 (6522) & 189 (2846) 
    \end{tabular}
\label{kidsAnnotationData}
\end{table}

\subsection{Pre-processing}
Through parameter optimization, we decided to use $32 \times 32$ pixel RGB images of face for training, validation and testing datasets. First, each video file was deconstructed into frames and all frames were fed into a cascade object detector implemented in MATLAB\textsuperscript{\textregistered} 9.8.0.1323502 (R2020a) based on Viola and Jones algorithm \cite{Viola01rapidobject}. Then, the face boundaries returned by the algorithms were used to crop each frame to only the face region. All the extracted frames for both neutral and tongue images were stored in separate folders. To avoid possible over-fitting, all the training and validation samples were augmented by randomly scaling and rotating, ranging from .5 to 1 and from -20 to 20 degrees, respectively.

\subsection{Network Architecture}
\begin{figure}[h]
\scalebox{.64}{
    \centering
        \begin{tikzpicture}[node distance=1.5cm]
        \node (in) [io] {Input [$32\times32$]};
        \node (conv1) [convolution, below of=in] {Convolution [$3\times3$]};
        \node (norm1) [normalize, below of=conv1] {Normalization};
        \node (relu1) [relu, below of=norm1] {ReLU};
        
        \node (pool1) [pooling, right of=relu1, xshift=2cm] {Max Pooling};
        \node (conv2) [convolution, above of=pool1] {Convolution [$3\times3$]};
        \node (norm2) [normalize, above of=conv2] {Normalization};
        \node (relu2) [relu, above of=norm2] {ReLU};
        
        \node (pool2) [pooling, right of=relu2, xshift=2cm] {Max Pooling};
        \node (conv3) [convolution, below of=pool2] {Convolution [$3\times3$]};
        \node (norm3) [normalize, below of=conv3] {Normalization};
        \node (relu3) [relu, below of=norm3] {ReLU};
        
        \node (dense) [dense, right of=relu3, xshift=2cm] {Fully Connected};
        \node (soft) [soft, above of=dense] {Soft Max};
        \node (class) [class, above of=soft] {Classification};
        \node (out) [io, above of=class] {Output};

        \draw [arrow] (in) -- (conv1);
        \draw [arrow] (conv1) -- (norm1);
        \draw [arrow] (norm1) -- (relu1);
        \draw [arrow] (relu1) -- (pool1);
        
        \draw [arrow] (pool1) -- (conv2);
        \draw [arrow] (conv2) -- (norm2);
        \draw [arrow] (norm2) -- (relu2);
        \draw [arrow] (relu2) -- (pool2);
        
        \draw [arrow] (pool2) -- (conv3);
        \draw [arrow] (conv3) -- (norm3);
        \draw [arrow] (norm3) -- (relu3);
        \draw [arrow] (relu3) -- (dense);
        
        \draw [arrow] (dense) -- (soft);
        \draw [arrow] (soft) -- (class);
        \draw [arrow] (class) -- (out);
        
        \end{tikzpicture}}
        \caption{Proposed network architecture for tongue detection algorithm}
    \label{networkArchitecture}
\end{figure}
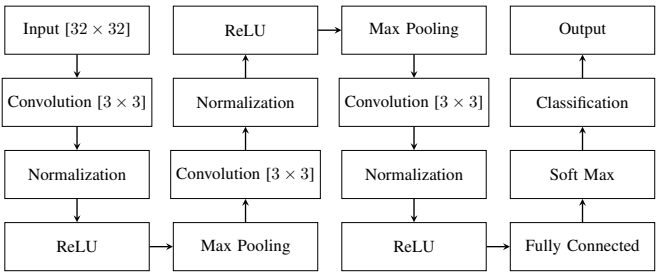

The proposed CNN architecture is illustrated in Figure \ref{networkArchitecture}. It consists of three convolution layers with a filter size of $3\times3$, zero-padding of 1 along all the edges, and 96, 32 and 64 filters, respectively. All max-pooling layers have a filter size of $2\times2$ with stride of 2. The max epoch and mini-batch size were set to 50 and 128, respectively. After the last rectified linear activation layer (ReLU), which outputs the input if it is positive and zero, otherwise, we used one fully connected layer, followed by softmax and classification layers. The training was performed using stochastic gradient descent with momentum optimizer.

\subsection{Hardware and software configurations}
We used a Pop!\_OS 18.04 LTS machine with Intel\textsuperscript{\textregistered} Core\texttrademark i7-8750H CPU, 24 GB of RAM, NVIDIA\textsuperscript{\textregistered} Quadro\textsuperscript{\textregistered} P1000 with 640 CUDA\textsuperscript{\textregistered} cores and 4 GB of GDDR5 RAM, using MATLAB\textsuperscript{\textregistered} 9.8.0.1323502 (R2020a) to run all the experiments.

\subsection{Experimental setup}\label{setup}
We implemented the method in MATLAB\textsuperscript{\textregistered} 9.8.0.1323502 (R2020a), and the code is available on GitHub\footnote{https://hollandbloorview.flintbox.com/?embed=true\#technologies/4996ecf2-3623-4dc1-a6e2-b54f302af48c}. To urge the importance of using children's dataset instead of adult datasets for pediatric research, we defined four different scenarios.

\subsubsection{Training with adults' data} In this case, we focused on adult's data and used it for training, and tested the generated model on the children dataset. The purpose of this scenario was to examine the possibility of using adult's data to train a network that can perform reasonably good on children's data. Since the adult's dataset included nose to chin images, we created a similar dataset of children's data to make both datasets as homogeneous as possible.

\subsubsection{Training with children's data} In this case, we only used children's data to train and test the network. The goal here was to provide practical evidence and emphasize the importance of having pediatric-specific datasets for AT applications.

\subsubsection{Training with a combination of adults' and children's data} In this case, we wanted to investigate the possibility of using both datasets to help generate a more generalizable model. Therefore, we combined both datasets and trained a network that can reasonably work in either situation.

\subsubsection{Training with adult's data and fine-tune using children's data} In this case, we wanted to improve the performance of the network trained on adult data and tested on children's images. We used transfer learning to modify and improve the network weights.

\section{Results}\label{sec:results}
We trained four networks corresponding to each scenario explained in Section \ref{setup}, and the results are reported in Table \ref{results} in terms of accuracy, sensitivity, specificity, F1-score, and precision.

\begin{table}[h]
\caption{The resulting classification accuracies, sensitivities, specificities and F1-scores of each proposing scenarios trained using the same network architecture}
\centering
\addtolength{\tabcolsep}{-2.8pt}
    \begin{tabular}{c c c c c c}
    \textbf{Scenario} & \textbf{Accuracy} & \textbf{Specificity} & \textbf{Sensitivity} & \textbf{F1-Score} & \textbf{Precision}\\\hline
    1 & 0.77$\pm$0.37 & 0.78$\pm$0.44 & 0.79$\pm$0.14 & 0.67$\pm$0.27 & 0.69$\pm$0.35\\  
    2 & 0.97$\pm$0.02 & 0.98$\pm$0.03 & 0.87$\pm$0.13 & 0.87$\pm$0.07 & 0.91$\pm$0.14\\
    3 & 0.80$\pm$0.38 & 0.78$\pm$0.44 & 0.95$\pm$0.07 & 0.78$\pm$0.32 & 0.75$\pm$0.37\\
    4 & 0.95$\pm$0.05 & 0.98$\pm$0.04 & 0.72$\pm$0.41 & 0.70$\pm$0.39 & 0.90$\pm$0.21\\\hline
    \end{tabular}
\label{results}
\end{table}

To further assess the performance of the trained models and their robustness in detecting tongue, we created a miscellaneous class of non-tongue protrusion by adding mouth opening and smiling movements to the neutral face images of the test dataset. The results are presented in Table \ref{extra}.

\begin{table}[h]
\caption{The resulting classification accuracies, sensitivities, specificities, F1-scores, and precisions after adding mouth opening and smiling samples to the test data}
\centering
\addtolength{\tabcolsep}{-2.8pt}
    \begin{tabular}{ c c c c c c}
    \textbf{Scenario} & \textbf{Accuracy} & \textbf{Specificity} & \textbf{Sensitivity} & \textbf{F1-Score} & \textbf{Precision}\\\hline
    1 & 0.63$\pm$0.32 & 0.63$\pm$0.37 & 0.72$\pm$0.23 & 0.30$\pm$0.11 & 0.22$\pm$0.11\\
    2 & 0.90$\pm$0.09 & 0.90$\pm$0.10 & 0.88$\pm$0.07 & 0.66$\pm$0.17 & 0.57$\pm$0.25\\
    3 & 0.69$\pm$0.35 & 0.66$\pm$0.38 & 0.96$\pm$0.05 & 0.45$\pm$0.22 & 0.32$\pm$0.19\\
    4 & 0.89$\pm$0.12 & 0.92$\pm$0.15 & 0.68$\pm$0.41 & 0.51$\pm$0.36 & 0.61$\pm$0.34\\\hline
    \end{tabular}
\label{extra}
\end{table}

To provide more insights on the performance of the proposed approach, Table \ref{scenario2} shows the participant-level results of scenario two after adding extra samples to the neutral face images.

\begin{table}[h]
\caption{The resulting classification accuracies, sensitivities, specificities, F1-scores, and precisions of each participant in Scenario 2 after adding mouth opening and smiling samples to the test data}
\centering
\addtolength{\tabcolsep}{-2.8pt}
    \begin{tabular}{ c c c c c c}
    \textbf{Participant} & \textbf{Accuracy} & \textbf{Specificity} & \textbf{Sensitivity} & \textbf{F1-Score} & \textbf{Precision}\\\hline
    C01 & 0.94 & 0.95 & 0.85 & 0.76 & 0.69\\
    C02 & 0.76 & 0.74 & 1.00 & 0.41 & 0.26\\
    C03 & 0.93 & 0.94 & 0.88 & 0.61 & 0.47\\
    C04 & 0.99 & 1.00 & 0.83 & 0.88 & 0.94\\
    C05 & 0.89 & 0.89 & 0.86 & 0.64 & 0.51\\\hline
    \end{tabular}
\label{scenario2}
\end{table}

\section{Discussion}\label{sec:discussion}
This work presents preliminary results on the automatic detection of tongue protrusion in children. Our proposed CNN architecture for tongue detection was validated on four different scenarios showing an average accuracy rate of up to 97.00\% in the identification of the tongue protrusion and up to 90.00\% when a miscellaneous class of non-tongue movements (e.g., neutral face, mouth opening, and smiling) was considered. Based on the resulting accuracies, the second scenario, where the network was trained and tested using only children data achieved the highest classification accuracy, specificity, and F1-score, but lower sensitivity rate. Such behaviour is caused by the amount of data for tongue-out gestures as compared to neutral face instances, for which providing further samples would improve the sensitivity further. In terms of precision, the second scenario reached the highest performance only in the first experiment (i.e., tongue-out vs. neutral face). 
The worst accuracy and F1-score were obtained with the network trained in scenario one where the adult data was solely used for training and the testing was against children's data. One notable item is the large standard deviation for accuracy, which is caused by some participants' quality of data. For example, participant C03 got bored during the experiments and performed the gestures doing abrupt and unpredictable movements throughout the facial movement repetitions. Combining both datasets in scenario three improved all the resulting measures compared to scenario one. The reason for a considerable improvement in sensitivity was due to providing more samples for the tongue-out gesture as a result of combining two datasets.
In scenario four, the network was trained using adult data, and the weights were modified and adjusted by retraining over children's data. Compared to scenarios one and three, where we trained the network using adults' data, and combination of adults' and children's datasets, we found improvements in the resulting accuracy and specificity, but not sensitivity. As we mentioned before, this is due to providing a smaller number of samples for modifying the network to train over children's tongue-out gestures.
To further assess the robustness of the trained networks, we added two more face gestures: smiling and mouth opening, to the neutral face sample set investigating their effect on all the performance measures. In other words, we wanted to examine how the output would be affected, if a participant used other face gestures than the ones the network was trained for. 
Based on the results, scenario two and one showed the highest and lowest classification accuracies and F1-scores, respectively. Although the specificity decreased for all scenarios, we found a small sensitivity increase for both scenarios two and three, which was caused by a decrease in the number of false-negative cases. Based on the F1-scores reported in Table \ref{results} and Table \ref{extra}, we noticed that the second scenario outperformed the other ones in detecting tongue-out gestures. This could be evidence that collecting data for pediatric populations is critical, as training networks with adults' data would not provide relatively accurate and robust models. This is important, particularly in cases where participants have severe movement disabilities such as cerebral palsy and other motor and brain disabilities.
The second experiment (i.e., tongue-out vs. non-tongue gestures) was performed to assess the robustness of the network. Our results showed a low rate of false-positive instances only for one participant (C04) whose precision reached 94.00\%. In fact, four out of five children reported precision rates below 70.00\%. False positives play a crucial role in developing algorithms for AT devices, as generating a high number of instances that wrongly activate the AT may increase the risk of abandonment. The poor performance could be improved by training the network to recognize four classes (i.e., tongue-out, smiling, mouth opening, and neutral face) instead of using only two classes (i.e., tongue-out vs. non-tongue-out movements). 
Lastly, our classification performance results are comparable to the results obtained by Sapaico et al. \cite{sapaico2011detection} that reached a 90.00\% average accuracy rate in 10 able-bodied adults detecting tongue-out gestures using normalized correlation matching. No previous studies attempted to identify tongue-out gestures against a miscellaneous class of mouth movements (i.e., smiling, mouth opening, and neutral face) using video analysis and deep learning techniques. Although this study is not the first attempt to use tongue gestures to control a laptop or an AT device \cite{niu2019tongue}, our research is the first developed for children. Niu et al. \cite{niu2019tongue} identified six distinct gesture categories using machine learning algorithms, obtaining an 83.00\% average accuracy rate with five able-bodied participants. Out of these six gesture categories, only two were the same that we used (i.e., mouth-close and tongue-down that correspond to our neutral face and tongue-out gestures, respectively). Niu et al. \cite{niu2019tongue} did not report the individual performance of the five participants for all the gesture categories but they indicated the average error rate in detecting the cursor movements for four classes (tongue-down, tongue-up, tongue-left, and tongue-right) and obtained 18.40\% for the tongue-down gesture (i.e. tongue-out) due mostly to the wrong gesture detection. Our deep learning approach seems to perform better than Niu and colleagues if we assume that they achieved an overall 80.00\% of accuracy for tongue-out detection.
Our findings showed for the first time in the literature that it is possible to detect tongue gestures in pediatric populations. As anticipated by Tai and colleagues \cite{tai2008review}, computer vision-based approaches for facial movements are exceptionally portable, and low-cost solutions considering the inexpensive USB web camera costs and their performance are comparable to sensor-based AT solutions. We also demonstrated that avant-garde deep-learning approaches achieved better performance in tongue gesture detection. Our results will allow researchers to explore possible pathways for new AT devices for children with severe physical disabilities.

\subsection{Limitations}
One major limitation of the proposed method is the unavailability of datasets for the pediatric populations. This has a significant effect on the amount of research being done in this area. Also, collecting data from children has greater barriers than adults due to stricter privacy concerns, which prevent authors from publishing these data. Moreover, sometimes children feel shy in performing face gestures, especially movements like the tongue protrusion. This has limited the number of samples that we were able to collect for this class of gestures. Some of the tongue-out repetitions were almost visually imperceptible, making it challenging to distinguish the tongue from the mouth. The proposed method is limited to the recognition of two classes (i.e. neutral face and tongue-out gesture), which makes it only applicable to binary tasks. Although the extension to multi-class models is possible through one-against-the-rest technique, collecting extended tongue gesture datasets would make the proposed method suitable for multi-class tongue gesture recognition applications.  
The lower performance obtained with C02 might be due to the naturalistic environment. We used a standardized position for the camera that was the same for all participants but children could move a bit around and change the chair position and the distance from the camera if they liked. This participant performed some movement repetitions farther than the 45cm provided by the data collection protocol. Also, C02 was particularly shy and did not like performing facial gestures in front of the researchers that collected the data. As we mentioned above, an additional issue that we noticed during the data collection, especially with younger children, is related to the possibility that participants get bored during the experiments because they have to repeat the same movements several times as happened with C03 and C05. This factor might have introduced extra variability to the network getting worse the quality of the training set.    

\section{Conclusions}\label{sec:conclude}
In this paper, we proposed a robust deep learning method to detect tongue protrusion in pediatric populations. This work shows how tongue gesture detection algorithms for children should be based on pediatric training datasets. Based on our preliminary results, a model trained on children may increase classification performance for children with CP and CCN. It is essential to evaluate our deep-learning approach with children with severe disabilities showing that tongue gestures can be used in building novel AT communication solutions.

\section{Acknowledgement}
This work was supported partially by the Cerebral Palsy Alliance Research Foundation, Mitacs through the Mitacs Elevate program, and Holland Bloorview Kids Rehabilitation Hospital Foundation.

\bibliographystyle{ieeetr}
\bibliography{mybib}

\end{document}